\documentclass{eeict}
\inputencoding{utf8}

\usepackage{amsmath}
\usepackage{amssymb}
\usepackage{float} 
\usepackage{subfig} 

\title{\uppercase{Gabor Wavelets in Image Processing}}
\author{David Bařina}
\programme{Doctoral Degree Programme (2), FIT BUT}
\emails{xbarin02@stud.fit.vutbr.cz}
\supervisor{Pavel Zemčík}
\emailv{zemcik@fit.vutbr.cz}

\abstract{
This work shows the use of a two-dimensional Gabor wavelets in image processing.
Convolution with such a two-dimensional wavelet can be separated into two series of one-dimensional ones.
The key idea of this work is to utilize a Gabor wavelet as a multiscale partial differential operator of a given order.
Gabor wavelets are used here to detect edges, corners and blobs.
A performance of such an interest point detector is compared to detectors utilizing a Haar wavelet and a derivative of a Gaussian function.
The proposed approach may be useful when a fast implementation of the Gabor transform is available or when the transform is already precomputed.
}

\keywords{Gabor function, wavelet, feature detection, interest point detection}

\begin{document}

\maketitle

\selectlanguage{english}

\section{Introduction}

A Gabor atom (or function) was proposed by Hungarian-born electrical engineer Dennis Gabor in 1946 \cite{Gabor1946}.
Nowadays, Gabor functions are frequently used for feature extraction, especially in texture-based image analysis (e.g., classification, segmentation or edge detection) and more practically in face recognition.
Many of image processing tasks can be seen in terms of a wavelet transform.
Informally speaking, the image can be seen under the lens with a magnification given by the scale of a wavelet.
In doing so, we can only see just the information that is determined by the shape of the used wavelet.
The Gabor atoms can also be seen in the words of a wavelet transform \cite{Mallat2009}.
Specifically, Gabor wavelets are created from one particular atom by dilation (and rotation in two-dimensional case).
These Gabor wavelets provide a complete image representation \cite{Lee1996}. 

In a two-dimensional case, the absolute square of a correlation between an image and the two-dimensional Gabor function provides a local spectral energy density concentrated around a given position and frequency in a certain direction. 
A two-dimensional convolution with a circular (non-elliptical) Gabor function is separable \cite{Loy2002} to series of one-dimensional ones.
J. G. Daugman discovered that simple cells in the visual cortex of mammalian brains can be modeled by Gabor functions \cite{Daugman1985}.
Thus, image analysis by the Gabor wavelets is similar to perception in the human visual system \cite{Lee1996}.

This work focuses on the use of Gabor wavelets in image processing algorithms, namely the interest point detection.
There are several approaches to the interest point detection using Gabor functions or wavelets.
More specifically, the two most common approaches involve the edge detection from the feature image \cite{Shao1994} or the corner detection using a combination of responses to several filters with a different orientation \cite{Quddus1998}.
In this paper, a new method based on the Gabor wavelets is proposed.
This approach differs from previous approaches mainly in the way the filter response is computed.
More specifically, the filter response is determined only in two perpendicular directions.
The nature of this approach consists in the use of responses to Gabor wavelets as the partial derivatives in the well known detectors (e.g., Canny edge detector, Harris corner detector, Hessian-based blob detector).
Such an approach may be useful when a fast implementation of the Gabor transform is available, e.g., \cite{Unser1994}, or when the transform is already precomputed.
My main contribution consists of the utilization of the Gabor wavelet as a multiscale partial differential operator.

\section{Gabor function}

In the one-dimensional case, the Gabor function consists of a complex exponential (a cosine or sine function, in real case) localized around $x=0$ by the envelope with a Gaussian window shape
\begin{equation}
	g_{\alpha,\xi}(x) = \sqrt{\alpha/\pi} \, e^{-\alpha x^2} \, e^{-i\xi x} ,
\end{equation}
for $\alpha \in \mathbb{R}^+$ and $\xi, x \in \mathbb{R}$, where $\alpha = (2\sigma^2)^{-1}$, $\sigma^2$ is a variance and $\xi$ is a frequency.
Dilation of the complex exponential function and shift of the Gaussian window when the dilation is fixed form kernel of a Gabor transform. 
The Gabor transform (a special case of the short-time Fourier transform) employs such kernel for time-frequency signal analysis.
The mentioned Gaussian window is the best time-frequency localization window in a sense of the Heisenberg uncertainty principle \cite{Mallat2009}.

In a two-dimensional case, the absolute square of the correlation between an image and a two-dimensional Gabor function provides the spectral energy density concentrated around a given position and frequency in a certain direction. 
Moreover, the two-dimensional convolution with a circular (non-elliptical) Gabor function is separable \cite{Loy2002} to series of one-dimensional ones
\begin{equation}
	g_{\alpha,\pmb{\xi}}(\pmb{x}) = g_{\alpha,\xi_0}(x_0) \, g_{\alpha,\xi_1}(x_1) ,
\end{equation}
for $\pmb{\xi} = (\xi_0,\xi_1)$ and $\pmb{x} = (x_0,x_1)$.
Here, the actual frequency of the two-dimensional function is determined by $\xi = (\xi_0^2 + \xi_1^2)^{1/2}$.
Furthermore, $\vartheta = \arctan(\xi_1/\xi_0)$ is an angle between $x$-axis and a line perpendicular to the ridges of a wave (wavefronts).

\begin{figure}[H]
	\centering

	\subfloat[]{
		\label{fig:gabor-lenna-3}
		\includegraphics[scale=0.25]{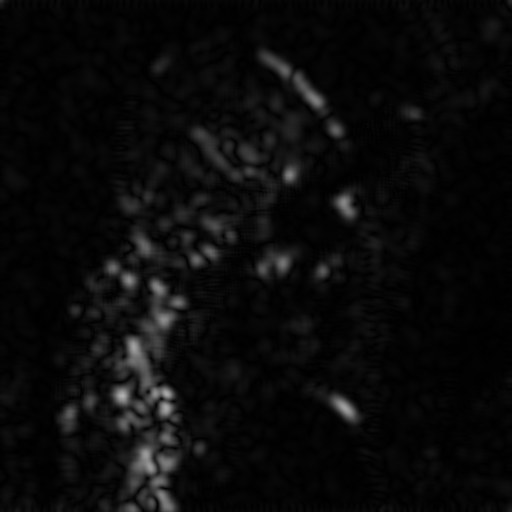}
	}
	\hspace{8pt}
	\subfloat[][]{
		\label{fig:lenna-gabor-max}
		\includegraphics[scale=0.25]{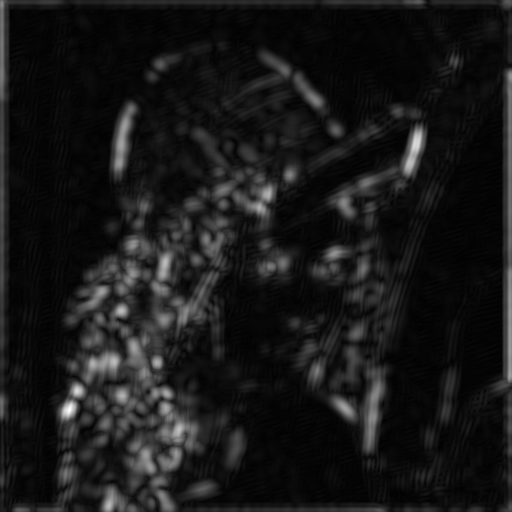}
	}
	\hspace{8pt}
	\subfloat[][]{
		\label{fig:gabor-filter}
		\includegraphics[scale=0.36]{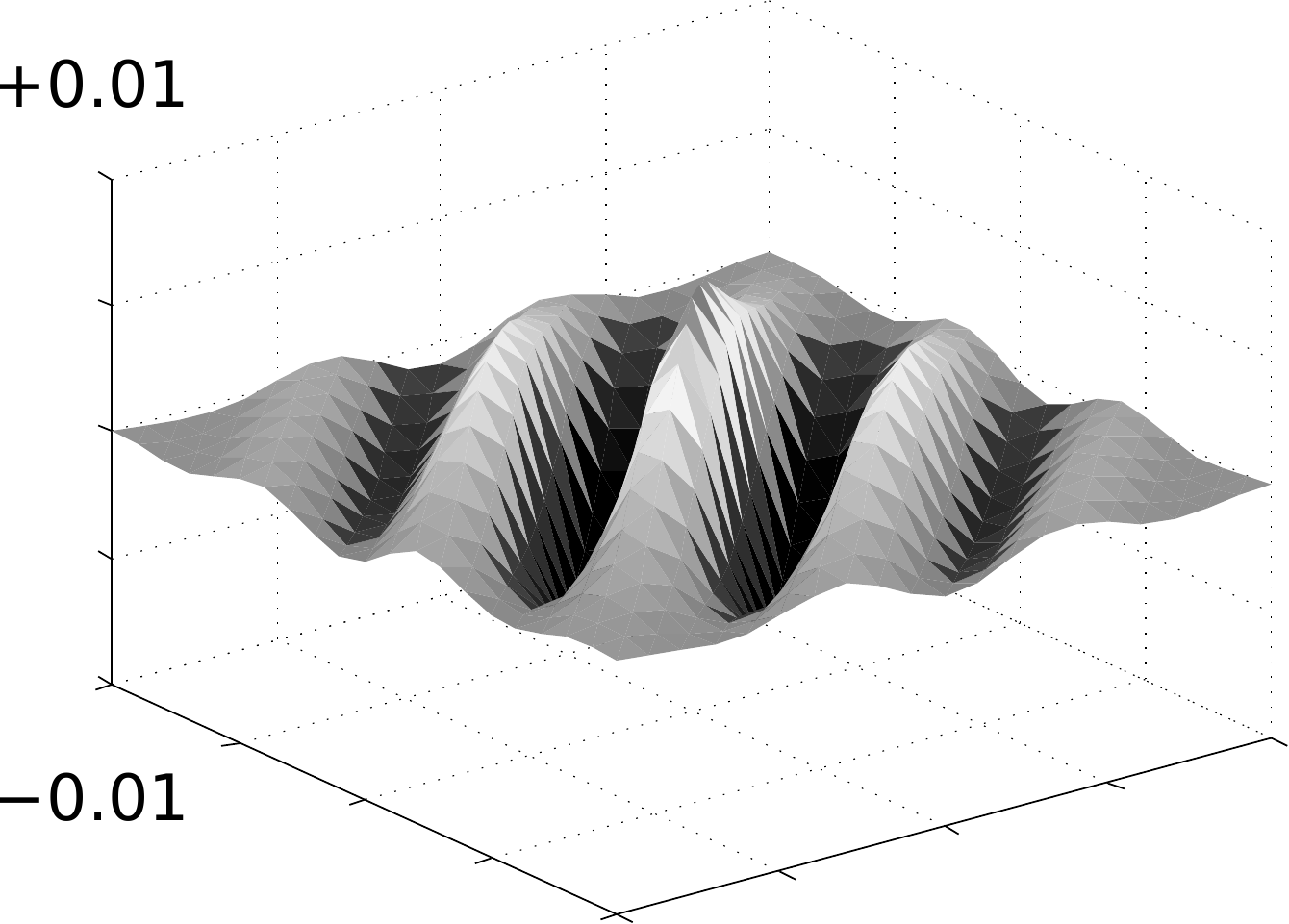}
	}

	\caption{\protect\subref{fig:gabor-lenna-3}~Gabor transform for direction of 137 degrees, used a kernel of size $11 \times 11$ pixels. \protect\subref{fig:lenna-gabor-max}~Gabor transform maxima for directions of 17, 77 and 137 degrees and \protect\subref{fig:gabor-filter}~one of the used Gabor functions.}
	\label{fig:lenna-gabor}
\end{figure}

As mentioned above, the Gabor function is often used to detect frequencies in various directions.
Figure \ref{fig:gabor-lenna-3} shows the Gabor transform for directions of 137 degrees.
Figure \ref{fig:lenna-gabor-max} shows the maxima of three partial transforms (for directions of 17, 77 and 137 degrees).
Figure \ref{fig:gabor-filter} shows one of the used Gabor functions.

%
%
%
%
%

\section{Gabor wavelet}

Elements of a family of mutually similar Gabor functions are called wavelets when they are created by dilation and shift from one elementary Gabor function (mother wavelet), i.e.
\begin{equation}
	g_{\alpha,\xi,a,b}(x) = |a|^{-1/2} \, g_{\alpha,\xi}\left(\frac{x-b}{a}\right)
\end{equation}
for $a \in \mathbb{R}^+$ (scale) and $b \in \mathbb{R}$ (shift).
By convention, the mother wavelet has the energy localized around $x=0$ as well as all of the wavelets are normalized $\|g\| = 1$.
Although the Gabor wavelets do not form orthonormal bases, the discrete set of them form a frame \cite{Lee1996}.

\section{Gabor wavelet as a differential operator}

An odd Gabor function (sine function) can be understood as a partial differential operator of an odd order, while an even Gabor function (cosine function) can be understood as a partial differential operator of an even order.
The $\alpha$ and $\xi$ parameters may be determined by an appropriate algorithm, for example, by minimizing the Euclidean distance from a differential operator we want to approximate.

\begin{figure}[H]
	\centering

	\subfloat[][]{
		\label{fig:gabor-sin}
		\includegraphics[scale=0.40]{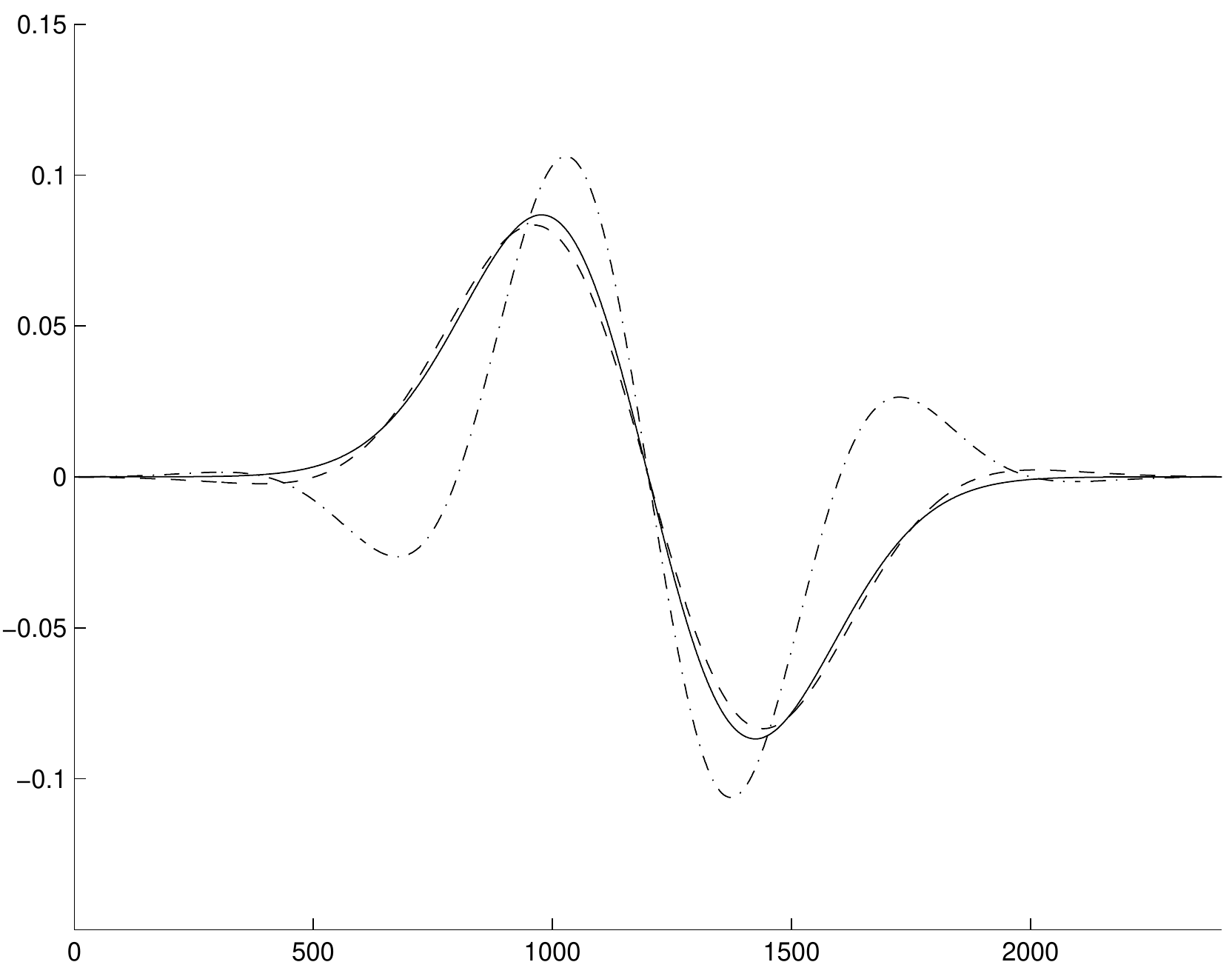}
	}
	\hspace{8pt}
	\subfloat[][]{
		\label{fig:gabor-cos}
		\includegraphics[scale=0.40]{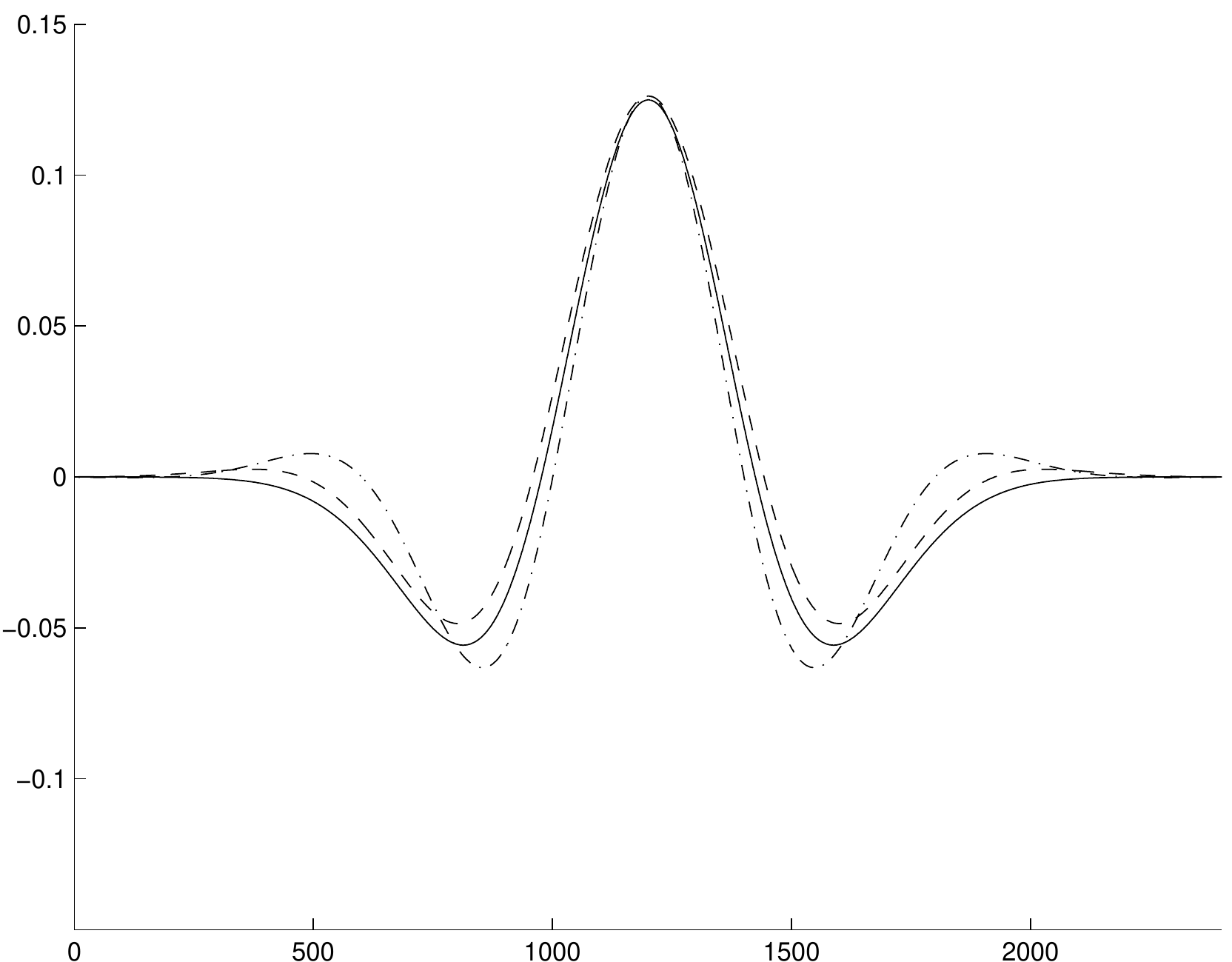}
	}

	\caption{
		Graphs of
		\protect\subref{fig:gabor-sin} the odd Gabor functions with $\xi=0.79$ (dash-dot), $\xi=0.45$ (dashed) and the first derivative of the Gaussian function (solid);
		\protect\subref{fig:gabor-cos} the even Gabor functions with $\xi=0.79$ (dash-dot), $\xi=0.65$ (dashed) and the second derivative of the Gaussian function (solid);
		all with $\alpha=0.05$.
	}
	\label{fig:gabor-matlab}
\end{figure}

It may be tempting to calculate the even and odd derivatives together with a single complex Gabor wavelet.
This way, we determine the odd derivative from an imaginary part of the function $\Im(g)$ and the even derivative from a real part of the function $\Re(g)$.
The situation is illustrated in Figure \ref{fig:gabor-matlab}.
Here, the derivatives of a Gaussian function are shown with a solid line.
Their good approximations by the Gabor functions are shown by a dashed line.
Finally, slightly worse approximations by the single complex Gabor function are shown by a dash-dot line (the odd and even functions have the same $\xi$ parameter).

\section{Applications}

The following are some applications in which derivatives can be obtained by using the Gabor wavelet.
The used notation is in accordance with \cite{Mikolajczyk2005}.
The first order partial derivative of image $I$ with respect to variable $x$ is denoted by $I_{x}$.
Analogously, $I_{xx}$ denotes the second order partial derivative with respect to $x$ and $I_{xy}$ is the second order mixed derivative.
Furthermore, $I_x(\pmb{x},\sigma_D)$ denotes a partial derivative obtained at the location of an point $\pmb{x}$ and calculated by using a Gabor wavelet with scale $a \varpropto \sigma_D$.

\subsection{Edge detection}

For the edge detection, the convolution in two perpendicular directions is performed with variously dilated wavelets (e.g., separately in row and column directions).
It is necessary to use a wavelet which serves as the first order partial differential operator (e.g., a first derivative of a Gaussian function \cite{Mallat2009}).
Consequently, local maxima of moduli
\begin{equation}
	M(\pmb{x},\sigma_D) =
		\sqrt{I_x^2(\pmb{x},\sigma_D) + I_y^2(\pmb{x},\sigma_D)}
\end{equation}
are found.
Only the maxima above a given threshold are considered (due to noise and slight edges).
As a result, the edges for each scale are obtained.

Another approach for the edge detection is to use a wavelet which serves as the second order partial differential operator.
In this case, edges are located in a zero-crossing points (sign changes, in the discrete case) \cite{Jan2006}.
This approach is used in detectors based on Laplacian, Laplacian of Gaussian (LoG) and Difference of Gaussians (DoG), see \cite{Jan2006}.

\subsection{Corner detection}

The key idea is to obtain the partial derivatives needed for a construction of an autocorrelation matrix (second moment matrix) \cite{Mikolajczyk2005}
\begin{equation}
	\mu(\pmb{x},\sigma_I,\sigma_D) = \sigma_D^2 \, g(\sigma_I) *
	\begin{bmatrix}
		I_x^2 (\pmb{x},\sigma_D) & I_xI_y(\pmb{x},\sigma_D) \\
		I_xI_y(\pmb{x},\sigma_D) & I_y^2 (\pmb{x},\sigma_D)
	\end{bmatrix}
\end{equation}
by using the convolution with the Gabor wavelets.
A Gaussian window of $\sigma_I$ scale is used for averaging of the derivatives.
On this matrix, e.g., Harris \& Stephens \cite{Harris1988}, Förstner and Shi-Tomasi detectors are based, see \cite{Kenney2005}.
Also here, it is necessary to use such a Gabor wavelet which serves as the first order partial differential operator.

\subsection{Blob detection}

Following the same principle, blobs can be detected \cite{Mikolajczyk2005} from the second order partial derivatives using a Hessian matrix
\begin{equation}
	H(\pmb{x},\sigma_D) =
	\begin{bmatrix}
		I_{xx}(\pmb{x},\sigma_D) & I_{xy}(\pmb{x},\sigma_D) \\
		I_{xy}(\pmb{x},\sigma_D) & I_{yy}(\pmb{x},\sigma_D)
	\end{bmatrix} .
\end{equation}

\begin{figure}[H]
	\centering

	\subfloat[][]{
		\label{fig:tk-lenna}
		\includegraphics[scale=0.25]{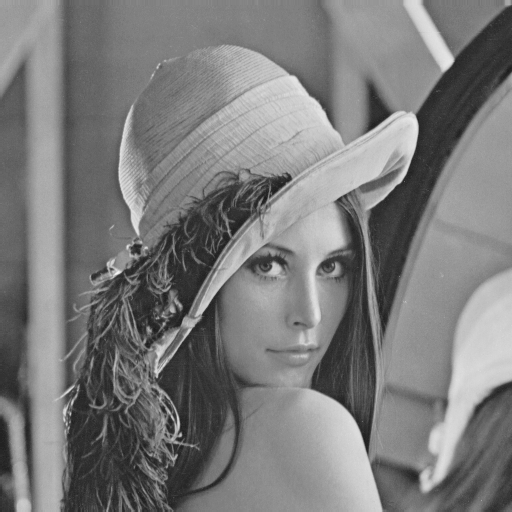}
	}
	\hspace{8pt}
	\subfloat[][]{
		\label{fig:tk-hessian}
		\includegraphics[scale=0.25]{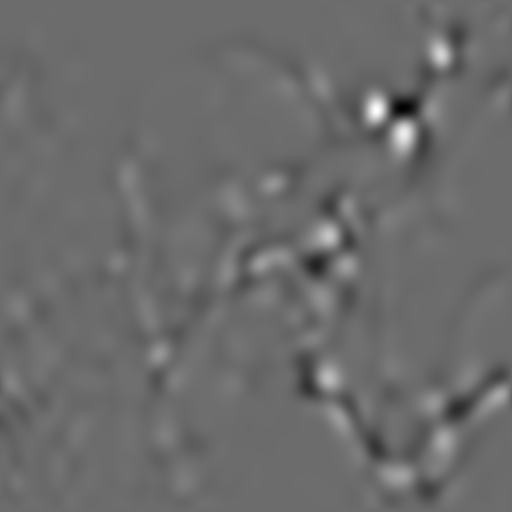}
	}
	\hspace{8pt}
	\subfloat[][]{
		\label{fig:tk-ellipses}
		\includegraphics[scale=0.25]{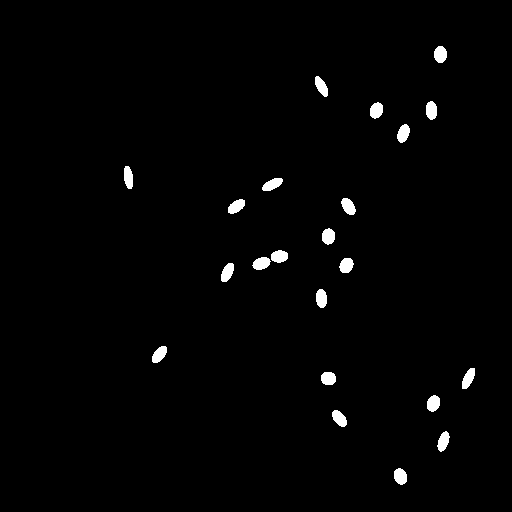}
	}

	\caption{
		Detection of affine covariant features:
		\protect\subref{fig:tk-lenna} input image,
		\protect\subref{fig:tk-hessian} determinant of the Hessian,
		\protect\subref{fig:tk-ellipses} the detected affine blobs.
	}
	\label{fig:tk-hesaff}
\end{figure}

An example of such a blob detection is shown in Figure~\ref{fig:tk-hesaff}.
Here, the complex Gabor wavelet serving as the first order partial differential operator has been used.

\section{Evaluation}

\begin{figure}[H]
	\centering

	\subfloat[][]{
		\label{fig:eval-1}
		\includegraphics[scale=0.40]{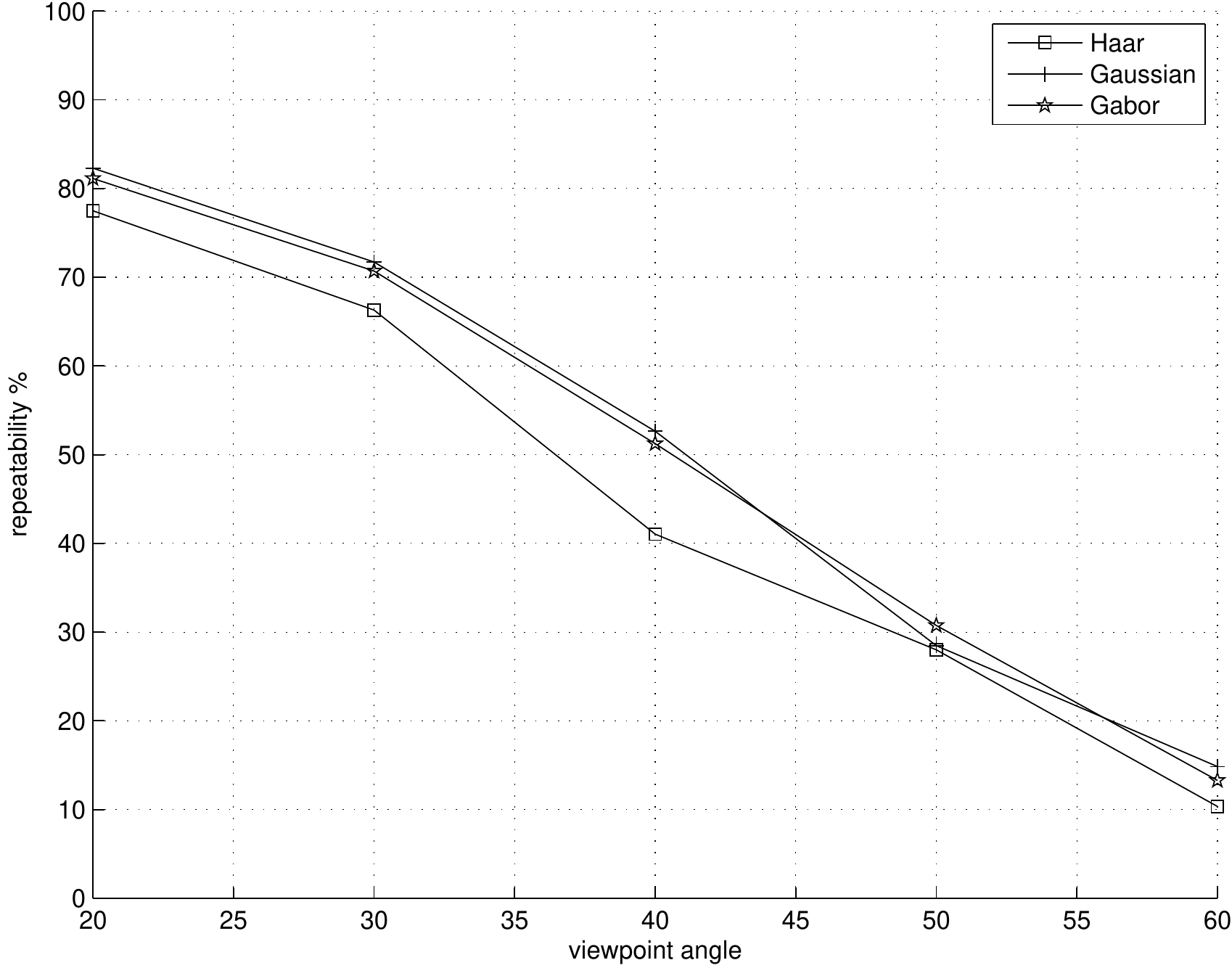}
	}
	\hspace{8pt}
	\subfloat[][]{
		\label{fig:eval-2}
		\includegraphics[scale=0.40]{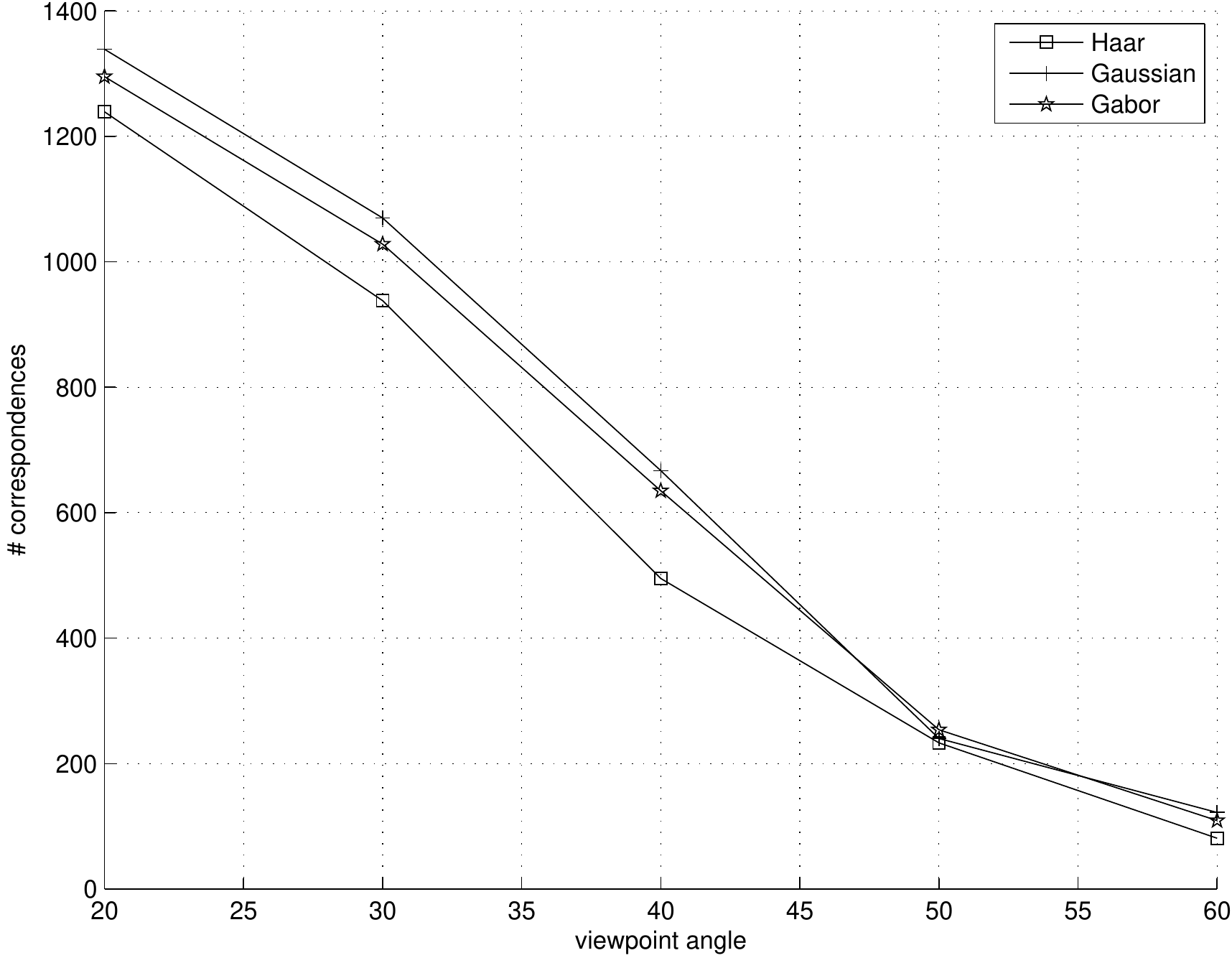}
	}

	\caption{
		Viewpoint change:
		\protect\subref{fig:eval-1} repeatability score and
		\protect\subref{fig:eval-2} number of corresponding regions.
	}
	\label{fig:eval}
\end{figure}

The proposed approach to the interest point detection has been evaluated by tools used by K.~Mi\-ko\-laj\-czyk in the evaluation of affine covariant region detectors \cite{Mikolajczyk2005}.
The performance is measured against changes in viewpoint for the structured scene.
The results of the evaluation are shown in Figure \ref{fig:eval}.
Here, three Hessian-based blob detectors was compared.
In these detectors, the partial derivatives was determined using the convolution with an appropriate derivative of the Gaussian function, the Haar wavelet and the Gabor wavelet.
The objective of this experiment is to measure the repeatability of the detectors, see \cite{Mikolajczyk2005}.
The best results are obtained with the derivative of the Gaussian function 
which is closely followed by the Gabor wavelet.
The worst results are achieved with the discontinuous Haar wavelet.

\section{Conclusion}

In this work, an outline how to use a two-dimensional separable Gabor wavelet to an interest point detection was proposed.
The main contribution of this work lies in the use of the Gabor wavelet as a multiscale partial differential operator.

\section*{Acknowledgment}

This work has been supported by the EU FP7-ARTEMIS project SMECY, grant no. 100230.

\bibliography{ref}

\begin{thebibliography}{10}

\bibitem{Daugman1985}
J.~G. Daugman.
\newblock Uncertainty relation for resolution in space, spatial frequency, and
  orientation optimized by two-dimensional visual cortical filters.
\newblock {\em JOSA A}, 2(7):1160--1169, July 1985.

\bibitem{Gabor1946}
D.~Gabor.
\newblock Theory of communication.
\newblock {\em Journal of the Institution of Electrical Engineers -- Part III:
  Radio and Communication Engineering}, 93(26):429--457, Nov. 1946.

\bibitem{Harris1988}
C.~Harris and M.~J. Stephens.
\newblock A combined corner and edge detector.
\newblock In {\em Alvey Vision Conference}, pages 147--152, 1988.

\bibitem{Jan2006}
J.~Jan.
\newblock {\em Medical Image Processing, Reconstruction and Restoration :
  Concepts and Methods}.
\newblock Signal Processing and Communications. Taylor \& Francis, 2006.

\bibitem{Kenney2005}
C.~S. Kenney, M.~Zuliani, and B.~S. Manjunath.
\newblock An axiomatic approach to corner detection.
\newblock {\em IEEE CVPR}, 1:191--197, 2005.

\bibitem{Lee1996}
T.~S. Lee.
\newblock Image representation using 2{D} {G}abor wavelets.
\newblock {\em IEEE Transactions on Pattern Analysis and Machine Intelligence},
  18(10):959--971, Oct. 1996.

\bibitem{Loy2002}
G.~Loy.
\newblock Fast computation of the {G}abor wavelet transform.
\newblock In {\em DICTA2002: Digital Image Computing Techniques and
  Applications}, pages 279--284, 2002.

\bibitem{Mallat2009}
S.~Mallat.
\newblock {\em A Wavelet Tour of Signal Processing : The Sparse Way. With
  contributions from Gabriel Peyré.}
\newblock Academic Press, 3 edition, 2009.

\bibitem{Mikolajczyk2005}
K.~Mikolajczyk, T.~Tuytelaars, C.~Schmid, A.~Zisserman, J.~Matas,
  F.~Schaffalitzky, T.~Kadir, and L.~Gool.
\newblock A comparison of affine region detectors.
\newblock {\em IJCV}, 65(1):43--72, 2005.

\bibitem{Quddus1998}
A.~Quddus and M.~M. Fahmy.
\newblock Corner detection using {G}abor-type filtering.
\newblock In {\em ISCAS '98: IEEE International Symposium on Circuits and
  Systems}, volume~4, pages 150--153, 1998.

\bibitem{Shao1994}
J.~Shao and W.~Förstner.
\newblock Gabor wavelets for texture edge extraction.
\newblock In {\em ISPRS Commission III Symposium on Spatial Information from
  Digital Photogrammetry and Computer Vision}, 1994.

\bibitem{Unser1994}
M.~Unser.
\newblock Fast {G}abor-like windowed {F}ourier and continuous wavelet
  transforms.
\newblock {\em IEEE Signal Processing Letters}, 1(5):76--79, May 1994.

\end{thebibliography}
\bibliographystyle{abbrv}

\end{document}